\documentclass{article}

\usepackage[preprint]{neurips_2025}

\usepackage[utf8]{inputenc}
\usepackage[T1]{fontenc}
\usepackage{hyperref}
\usepackage{url}
\usepackage{booktabs}
\usepackage{amsfonts}
\usepackage{amsmath}
\usepackage{nicefrac}
\usepackage{microtype}
\usepackage{xcolor}
\usepackage{graphicx}
\usepackage{algorithm}
\usepackage{algorithmic}
\usepackage{multirow}
\usepackage{subcaption}
\usepackage{enumitem}
\usepackage{tikz}
\usetikzlibrary{arrows.meta, positioning, shapes.geometric, fit, calc, backgrounds}
\usepackage{amsthm}
\newtheorem{definition}{Definition}

\title{AI-Supervisor: Autonomous AI Research Supervision\\via a Persistent Research World Model}

\author{
  Yunbo Long \\
  \texttt{longyunbo218@gmail.com}
}

\begin{document}

\maketitle

\begin{abstract}
Existing automated research systems operate as stateless, linear pipelines --- generating outputs without maintaining any persistent understanding of the research landscape they navigate. They process papers sequentially, propose ideas without structured gap analysis, and lack mechanisms for agents to verify, challenge, or refine each other's findings. We present \textbf{AI-Supervisor}, a multi-agent orchestration framework where specialized agents provide end-to-end AI research supervision driven by human interests --- from literature review through gap discovery, method development, evaluation, and paper writing --- through autonomous exploration and self-correcting updates of research knowledge. Unlike sequential pipelines, AI-Supervisor maintains a continuously evolving \emph{Research World Model}, implemented as a Knowledge Graph, that captures methods, benchmarks, known limitations, and unexplored gaps, serving as shared memory across all agents and enabling agents to explore and build upon a structured understanding of the research landscape. The framework introduces three architectural contributions: (1) \emph{structured gap discovery} that decomposes methods into core modules, validates their performance across benchmarks, and maps the specific gaps each module creates; (2) \emph{self-correcting discovery loops} that probe why modules succeed on certain problems and fail on others, whether benchmarks carry hidden biases, and whether evaluation protocols remain adequate for emerging challenges; and (3) \emph{self-improving development loops} governed by cross-domain mechanism search that iteratively targets failing modules by finding solutions from other scientific fields. All agents operate under a \emph{consensus mechanism} where independent findings are corroborated before being committed to the Research World Model. The framework is model-agnostic, supports all mainstream large language models, and scales elastically with token budget --- from lightweight exploration to full-scale investigation. Code is available at \url{https://github.com/autoproflab-debug/AI-Supervisor}.
\end{abstract}

\section{Introduction}
\label{sec:intro}

\begin{center}
\textit{``I have no special talents. I am only passionately curious.''}\\
\vspace{0.1cm}
\begin{flushright}
--- Albert Einstein, 1952
\end{flushright}
\end{center}

\begin{figure}[t]
\centering
\includegraphics[width=\textwidth]{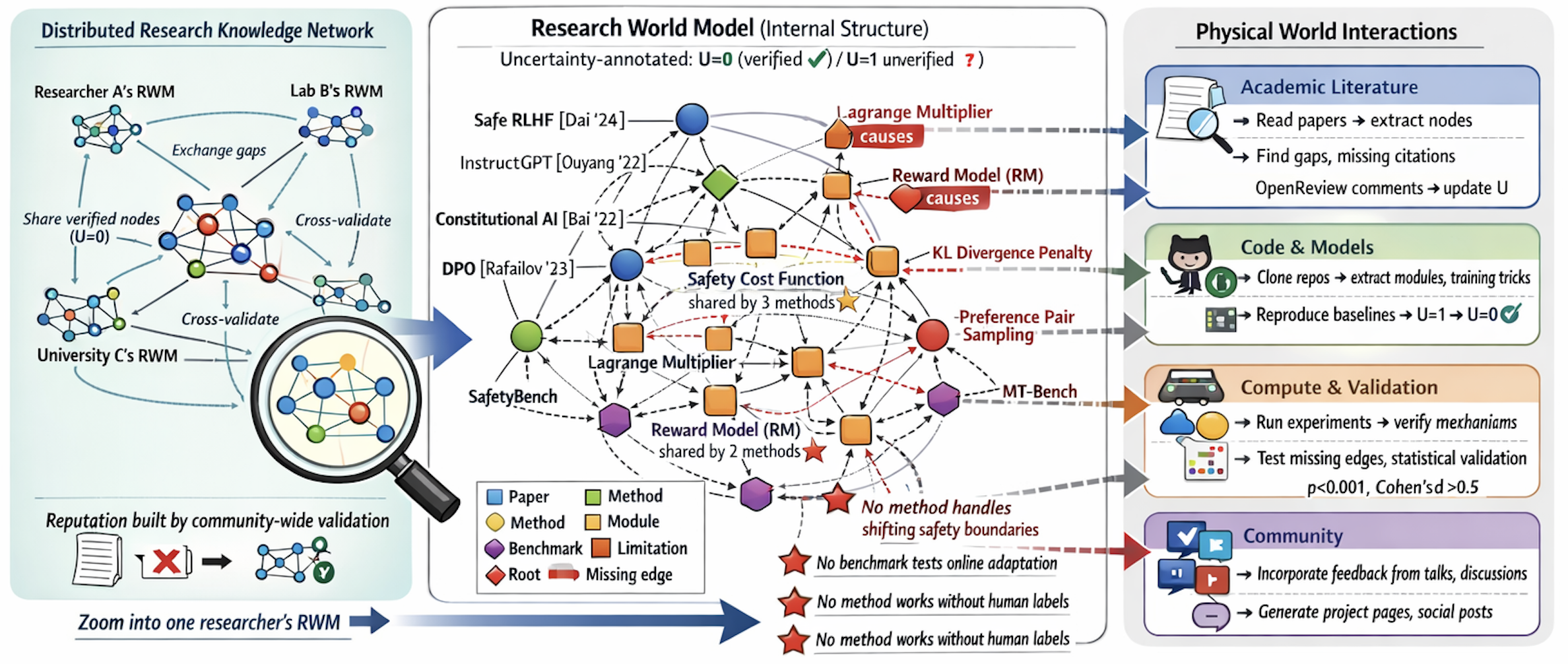}
\caption{The Research World Model ecosystem. Left: distributed knowledge network of multiple RWMs sharing verified findings. Center: internal structure of one RWM with uncertainty-annotated edges carrying performance metrics. Right: bidirectional interaction with physical world—literature, code, compute, and community—enabling active exploration over passive generation.}
\label{fig:examples}
\end{figure}

\noindent Today, virtually all AI research is driven by project funding or corporate sponsorship rather than personal curiosity. This is because research supervision --- the intellectual guidance needed to navigate literature, identify gaps, design experiments, and survive peer review --- remains controlled by a small number of universities and advanced companies, and gaining access to this supervision requires institutional affiliation. As a consequence, AI research cannot be personalized: an individual who wants to pursue their own research interest has no way to obtain a professional supervisor and lab to support their personal project, unless that interest happens to align with an existing funded program. This means that most AI research directions, publications, and applications are ultimately determined by the priorities of a few institutions, not by the broader community's curiosity. The problem persists even \emph{inside} these institutions: faculty are overwhelmed, research groups are oversubscribed, and many PhD and Master's students cannot secure funded positions because funding comes with fixed topics chosen by the supervisor --- leaving students to either abandon their interests or navigate research alone without guidance.
This concentration is reinforced by prohibitive research sources~\citep{cottier2024costs}, closed-source frontier models~\citep{nature2025opensource, fradkin2025pricing}, an academic job market where fewer than 30\% of PhD graduates secure permanent positions~\citep{stanfordai2025, kwon2025phd}, and a publication system that structurally favors large teams over individual researchers~\citep{nature2025reviews, icml2025crisis} --- widening global inequality in AI access~\citep{undp2025divergence, imf2025inequality}. The solution is not to give everyone access to more compute --- it is to give every individual their own AI research team.

Besides, we argue for a new era in which AI research follows \emph{personal interests}. As more and more people use AI and seek to do AI research --- whether for applications, publications, or pure curiosity --- the traditional model of human supervision by a small number of professors and corporate research leads cannot scale to meet this demand. But recent advances in large language models and agentic AI make it potentially possible for \emph{anyone} to access world-leading research supervision without attending a top institute or joining an elite lab --- to have a professional AI research team that reads literature, discovers gaps, develops methods, and writes papers \emph{for their own chosen topic}. This paper argues that AI research supervision itself can be automated, enabling curiosity-driven, personalized research at scale.
Systems like the AI Scientist~\citep{lu2024aiscientist, lu2025aiscientistv2}, AI-Researcher~\citep{tang2025airesearcher}, and Agent Laboratory~\citep{schmidgall2025agentlab} demonstrate that LLM agents can automate portions of the AI research pipeline --- generating ideas, running experiments, writing papers. However, these systems still assume an experienced researcher at the helm: someone who knows which problems matter, where the gaps are, and how to evaluate rigor. They automate the \emph{execution} of research while leaving the hardest part --- \emph{research supervision} --- to humans. Fundamentally, almost all existing methods treat automated research as a \emph{generation} task --- using existing knowledge to prompt LLMs to produce new text --- rather than as \emph{active exploration and interaction} with a research knowledge world. We argue that when a research interest is provided, LLM agents should actively interact with real-world research knowledge to construct new understanding, not merely generate plausible-sounding text. This means validating claims through actual computation on GPUs and APIs, engaging with the broader research community by incorporating reviewer feedback from platforms like OpenReview, and --- crucially --- maintaining and continuously updating a \emph{persistent research world model} throughout the exploration process, so that every discovery, verification, and failure is recorded and informs the next step. This entire cycle of exploration, validation, and world-model maintenance is how new knowledge is genuinely created, as opposed to merely generated. Yet no existing system operates this way. Meanwhile, evidence suggests that AI tools, while expanding individual productivity, may actually \emph{narrow} the collective scope of research~\citep{hao2026aitools} --- making the shift from passive generation to active exploration all the more urgent.

We present AI-Supervisor, a framework that automates \emph{AI research supervision itself} through self-correcting multi-agent consensus on a \emph{Persistent Research World Model}. AI-Supervisor takes a user's research interest --- stated in plain language, with no domain expertise required --- and provides the full intellectual scaffolding that a world-class supervisor would: reading literature, identifying what the field is missing, searching across domains for solutions, testing hypotheses with statistical rigor, and iterating until the contribution meets publication standards. The framework is model-agnostic, supporting all mainstream LLMs worldwide (GPT-4, Claude, Gemini, LLaMA, Qwen, DeepSeek, and others). The user brings their curiosity; AI-Supervisor brings the lab, the methodology, and the team. Figure~\ref{fig:examples} illustrates the range of AI research scenarios that AI-Supervisor can handle --- from a student's first research question to cross-domain method development across diverse fields.
We evaluate AI-Supervisor through case studies across multiple AI research domains. Our contributions are:

\begin{itemize}[leftmargin=*,nosep]
    \item \textbf{Persistent Research World Model.} We introduce the first research automation system built around a continuously evolving world model of the research landscape, implemented as an uncertainty-annotated Knowledge Graph. The Research World Model captures methods, modules, benchmarks, gaps, and limitations with typed edges and uncertainty states ($U{=}0$ verified, $U{=}1$ unverified), serving as shared memory, orchestration backbone, and quality-control mechanism across all agents. Unlike stateless pipelines, the Research World Model grows across sessions and projects, enabling structural gap reasoning and cross-project knowledge transfer.

    \item \textbf{Self-correcting multi-agent consensus.} We design a probing protocol where parallel agents independently investigate methods, benchmarks, and assumptions, then share all findings for cross-verification. An orchestrator aggregates collective evidence to produce verified gaps --- replacing the speculative gap identification of prior systems with empirically grounded discovery. Only findings corroborated across multiple agents are committed to the Research World Model.

    \item \textbf{Cross-domain self-improving development loops.} We propose a mechanism-first approach: root-cause analysis maps domain-specific failures to abstract problems, enabling search across \emph{other} scientific fields for solutions. A quality-gated checklist governs iteration with deterministic routing back to direction reassessment --- not just more searching --- when criteria fail. The Research World Model tracks what has been searched, what worked, and what failed, preventing duplicate effort across iterations.

    \item \textbf{Open-source, model-agnostic framework.} We release AI-Supervisor as composable skills compatible with all mainstream LLMs, designed to scale elastically with token budget.\footnote{Available at: \url{https://github.com/autoproflab-debug/AI-Supervisor}}
\end{itemize}

\section{Related Work}
\label{sec:related}

\subsection{End-to-End Research Automation}

We compare existing systems along five core pipeline stages: \emph{literature review}, \emph{reproduction \& validation}, \emph{gap analysis}, \emph{method development}, and \emph{evaluation}.

\textbf{Literature.} The AI Scientist v1~\citep{lu2024aiscientist} uses Semantic Scholar only for novelty checking (verifying whether a generated idea already exists), not systematic survey; v2~\citep{lu2025aiscientistv2} integrates literature queries into the idea generation loop but still for novelty filtering rather than field mapping. AI-Researcher~\citep{tang2025airesearcher} retrieves 10--15 reference papers from arXiv with GitHub code filtering and extracts mathematical formulations via RAG, but does not search across venues or extract reviewer weaknesses. Agent Laboratory~\citep{schmidgall2025agentlab} queries arXiv iteratively (top 20 abstracts per query), but this is the most failure-prone phase (60--80\% success rate). MLR-Copilot~\citep{du2024mlrcopilot} retrieves papers via Semantic Scholar for a single input paper's gaps. PaperQA2~\citep{lala2025paperqa} and OpenScholar~\citep{asai2026openscholar} provide strong literature synthesis but cover only this single stage. None perform multi-venue parallel search with reviewer score extraction. \emph{AI-Supervisor addresses this} by launching parallel search agents across 6--12 venues simultaneously, extracting OpenReview scores and reviewer weaknesses as community-validated gap signals, and building a ranked literature base through two-pass scoring (abstract filtering then full-paper deep reading).

\textbf{Reproduction.} No existing system independently reproduces baselines. AI Scientist v1 starts from human-authored code templates (NanoGPT, 2D Diffusion, Grokking) with pre-run baselines; v2 generates code from scratch but does not validate reported numbers from prior work. AI-Researcher analyzes existing implementations via bidirectional theory-code mappings but does not re-execute them. MLR-Copilot retrieves ``prototype code'' as a starting point. Agent Laboratory and ResearchAgent~\citep{baek2025researchagent} perform no reproduction at all. Without reproduction, no system can verify whether reported claims hold before building on them. \emph{AI-Supervisor addresses this} by cloning the top 5 methods into a unified evaluation repository, auto-detecting available compute, and reproducing each method on its own benchmarks --- updating KG edges to verified ($U=0$) or failed ($U=1$) before any gap analysis begins.

\textbf{Gap analysis.} AI Scientist v1 generates ideas as incremental modifications of code templates constrained by seed ideas; v2 uses open-ended LLM brainstorming ($\sim$20 ideas per prompt, human-selected to $\sim$3). AI-Researcher employs a divergent-convergent framework: generating five distinct directions then filtering by novelty, soundness, and transformative potential. MLR-Copilot fine-tunes a Llama3-7B IdeaAgent with RL (reward models for novelty, feasibility, effectiveness). ResearchAgent augments LLM ideation with an entity co-occurrence matrix that enables cross-domain connections (e.g., linking CRISPR with genetic reference panels). Agent Laboratory requires the human to provide the research idea entirely. Critically, \emph{all} of these approaches generate gap hypotheses from text analysis or LLM reasoning --- none probe where existing methods actually fail through empirical testing. \emph{AI-Supervisor addresses this} by deploying parallel probing agents that empirically test methods, benchmarks, and assumptions --- running actual cross-benchmark experiments to discover where methods fail, with an orchestrator achieving consensus across agents before committing verified gaps to the KG.

\textbf{Method development.} AI Scientist v1 uses linear code editing via Aider; v2 introduces progressive agentic tree search across four stages (preliminary investigation, hyperparameter tuning, research agenda, ablation) with VLM feedback on figure quality --- a major advance but still without cross-domain search. AI-Researcher uses cyclic development with explicit quality gates in Docker containers, where an Advisor Agent reviews against ``atomic concepts.'' Agent Laboratory's \texttt{mle-solver} samples from top-performing programs with LLM-scored fitness. MLR-Copilot's ExperimentAgent modifies retrieved prototype code. ResearchAgent and SciAgents~\citep{ghafarollahi2025sciagents} produce text proposals only --- no executable code. None search other scientific fields for techniques addressing the underlying mechanism of a gap. \emph{AI-Supervisor addresses this} through 5-WHY root-cause analysis that maps failing modules to abstract mechanisms, then searches other scientific fields using their vocabulary --- with a quality-gated checklist that routes back to direction reassessment (not just deeper search) when criteria fail.

\textbf{Evaluation.} AI Scientist v1 uses an ensemble of 5 LLM reviewers; v2 adds VLM-based figure critique and achieved one ICLR workshop acceptance (scores 6/7/6). AI-Researcher performs two-stage evaluation: code review (static analysis + runtime verification) followed by pairwise comparison against ground-truth papers using multiple LLMs. Agent Laboratory uses both human evaluation (10 PhD students) and NeurIPS-style automated reviewers, finding that automated reviewers overestimate quality (6.1/10 vs. human 3.8/10). No system performs cross-benchmark evaluation, ablation with statistical significance across seeds, or tests on benchmarks the method was \emph{not} designed for. \emph{AI-Supervisor addresses this} with multi-seed evaluation (3 seeds for mean$\pm$std), cross-model generalization testing, component ablation, and qualitative error analysis --- followed by an automated review loop that diagnoses each weakness and routes to the correct pipeline stage for fixing.

\subsection{Knowledge Graphs and World Models for Scientific Discovery}

Knowledge graphs have a long history in organizing scientific knowledge~\citep{hogan2021knowledge}, but their use as \emph{orchestration backbones} for automated research is recent. SciAgents~\citep{ghafarollahi2025sciagents} is the closest to AI-Supervisor in this regard: it pre-constructs a large-scale ontological KG (stored as GraphML with semantic embeddings) from scientific papers in bio-inspired materials, then uses graph-topology-driven gap finding --- sampling random paths within the KG to discover research opportunities at concept intersections. Four specialized agents (Ontologist, Scientist 1, Scientist 2, Critic) collaboratively generate $\sim$8,100-word research proposals from these graph paths. This demonstrates the power of KG-grounded reasoning: gaps emerge from graph structure rather than LLM hallucination. However, SciAgents' KG is domain-specific and static (pre-built, not updated during the research process), the system produces text proposals only (no executable code), and there are no reproduction, evaluation, or method development stages --- making it a discovery tool, not a research pipeline.

ResearchAgent~\citep{baek2025researchagent} takes a different approach: an entity co-occurrence matrix ($m \times m$ sparse matrix from BLINK entity linking across $\sim$50,000 entities) captures cross-domain associations, enabling idea augmentation beyond the immediate field --- for example, connecting ``Drosophila Genetic Reference Panel'' with ``CRISPR'' through co-occurrence patterns. Fifteen ReviewingAgents provide iterative feedback over $\sim$3 refinement rounds. However, this is a flat co-occurrence structure, not a typed KG with uncertainty annotations, and the system stops at idea generation without implementation or verification. KARMA~\citep{karma2025} uses nine collaborative agents for KG enrichment (entity discovery, relation extraction, schema alignment) achieving 83.1\% correctness on PubMed articles, but is a construction tool, not a research system.

\emph{AI-Supervisor's Research World Model} differs from these knowledge structures in four ways: (1) it is \emph{built during} the research process through section-specific extraction by parallel agents, not pre-constructed --- it evolves as the research progresses; (2) both nodes and edges carry \emph{uncertainty annotations} ($U \in \{0, 1\}$) --- every node starts unverified ($U=1$) and is promoted to verified ($U=0$) only after empirical testing, while every edge carries actual performance metrics, so the world model encodes not just \emph{what is claimed} but \emph{whether it holds} and \emph{how well}; (3) edges are actively validated during probing --- when a method's reported performance cannot be reproduced, the edge is flagged as $U=1$, making discrepancies visible for gap analysis; and (4) it serves as the \emph{orchestration backbone} --- routing decisions, gap discovery, and cross-session persistence all flow through the Research World Model rather than through conversation history or flat state. In this sense, AI-Supervisor's agents do not merely process papers --- they build and maintain a \emph{persistent world model} of the research landscape that grows smarter with each project.

\subsection{Multi-Agent Orchestration and Consensus for Scientific Discovery}

Multi-agent systems have emerged as a powerful paradigm for complex reasoning tasks~\citep{wang2024survey}. In research automation, agent architectures range from single-model pipelines (AI Scientist v1) to role-based teams (Agent Laboratory's PhD/Postdoc/Professor hierarchy) to specialized agent networks (AI-Researcher's 9 agents with mentor-student dynamics). AutoGen~\citep{wu2023autogen} and MetaGPT~\citep{hong2024metagpt} provide general-purpose multi-agent frameworks with event-driven architectures, but these offer infrastructure without research-domain knowledge or quality-gated iteration.

A critical open question is how multi-agent systems achieve \emph{consensus} --- how independent agents with potentially conflicting findings converge on reliable conclusions. In Agent Laboratory, consensus is implicit: agents follow a fixed sequential pipeline where each role's output feeds the next. In AI-Researcher, an Advisor Agent reviews against atomic concepts, but the review is unilateral rather than collective. In SciAgents, the Critic agent provides feedback but the interaction is sequential (Ontologist $\to$ Scientist 1 $\to$ Scientist 2 $\to$ Critic), not parallel with shared visibility.

\emph{AI-Supervisor introduces an explicit consensus mechanism}: parallel agents independently investigate distinct research questions, then \emph{all} agents see \emph{all} results before proposing next steps. The orchestrator aggregates collective evidence --- merging complementary discoveries, terminating unproductive lines, and redirecting effort --- so that routing decisions reflect the team's shared knowledge rather than any single agent's judgment. Only findings corroborated across multiple agents or verified by empirical testing are committed to the KG with $U=0$. This design prevents the single-point-of-failure problem inherent in sequential architectures, where one agent's error propagates uncorrected through the entire pipeline.

\vspace{0.5em}
\noindent\textbf{Summary.} Table~\ref{tab:comparison} compares systems across key capabilities. Several systems offer partial coverage: AI Scientist v2 provides iterative tree search with quality gates (Self-Imp.), AI-Researcher adds systematic literature retrieval and cyclic development, Agent Lab and ResearchAgent include iterative refinement with reviewer feedback, and SciAgents uses a pre-built ontological KG. However, no prior system combines curiosity-driven initiation, empirical gap probing, cross-domain search, a persistent and evolving world model with uncertainty annotations, and multi-agent consensus.

\begin{table}[t]
\caption{Comparison of AI-Supervisor with existing systems. \textbf{Curiosity}: starts from user interest, no expertise needed. \textbf{Lit}: systematic literature search. \textbf{Repro}: baseline reproduction. \textbf{Gap}: empirically verified gap probing. \textbf{Cross}: cross-domain search. \textbf{Self-Imp.}: self-correcting quality-gated loops. \textbf{RWM}: persistent evolving Research World Model with uncertainty. \textbf{Cons.}: multi-agent consensus. $^\dagger$Static pre-built KG, not persistent or evolving.}
\label{tab:comparison}
\centering
\scriptsize
\resizebox{\textwidth}{!}{%
\begin{tabular}{lcccccccc}
\toprule
\multirow{2}{*}{\textbf{System}} & \textbf{Curiosity} & & & \textbf{Gap} & \textbf{Cross-} & \textbf{Self-} & \multirow{2}{*}{\textbf{RWM}} & \multirow{2}{*}{\textbf{Cons.}} \\
& \textbf{Driven} & \textbf{Lit} & \textbf{Repro} & \textbf{Probing} & \textbf{Domain} & \textbf{Imp.} & & \\
\midrule
AI Scientist v2 & \texttimes & \texttimes & \texttimes & \texttimes & \texttimes & \checkmark & \texttimes & \texttimes \\
AI-Researcher & \texttimes & \checkmark & \texttimes & \texttimes & \texttimes & \checkmark & \texttimes & \texttimes \\
Agent Lab & \texttimes & \texttimes & \texttimes & \texttimes & \texttimes & \checkmark & \texttimes & \texttimes \\
SciAgents & \texttimes & \texttimes & \texttimes & \texttimes & \texttimes & \texttimes & $\dagger$ & \texttimes \\
ResearchAgent & \texttimes & \checkmark & \texttimes & \texttimes & \texttimes & \checkmark & \texttimes & \texttimes \\
PaperQA2 & \texttimes & \checkmark & \texttimes & \texttimes & \texttimes & \texttimes & \texttimes & \texttimes \\
MLR-Copilot & \texttimes & \texttimes & \texttimes & \texttimes & \texttimes & \texttimes & \texttimes & \texttimes \\
\midrule
\textbf{AI-Supervisor} & \checkmark & \checkmark & \checkmark & \checkmark & \checkmark & \checkmark & \checkmark & \checkmark \\
\bottomrule
\end{tabular}}
\end{table}

\section{AI-Supervisor Framework}
\label{sec:method}

\begin{figure}[t]
\centering
\includegraphics[width=\textwidth]{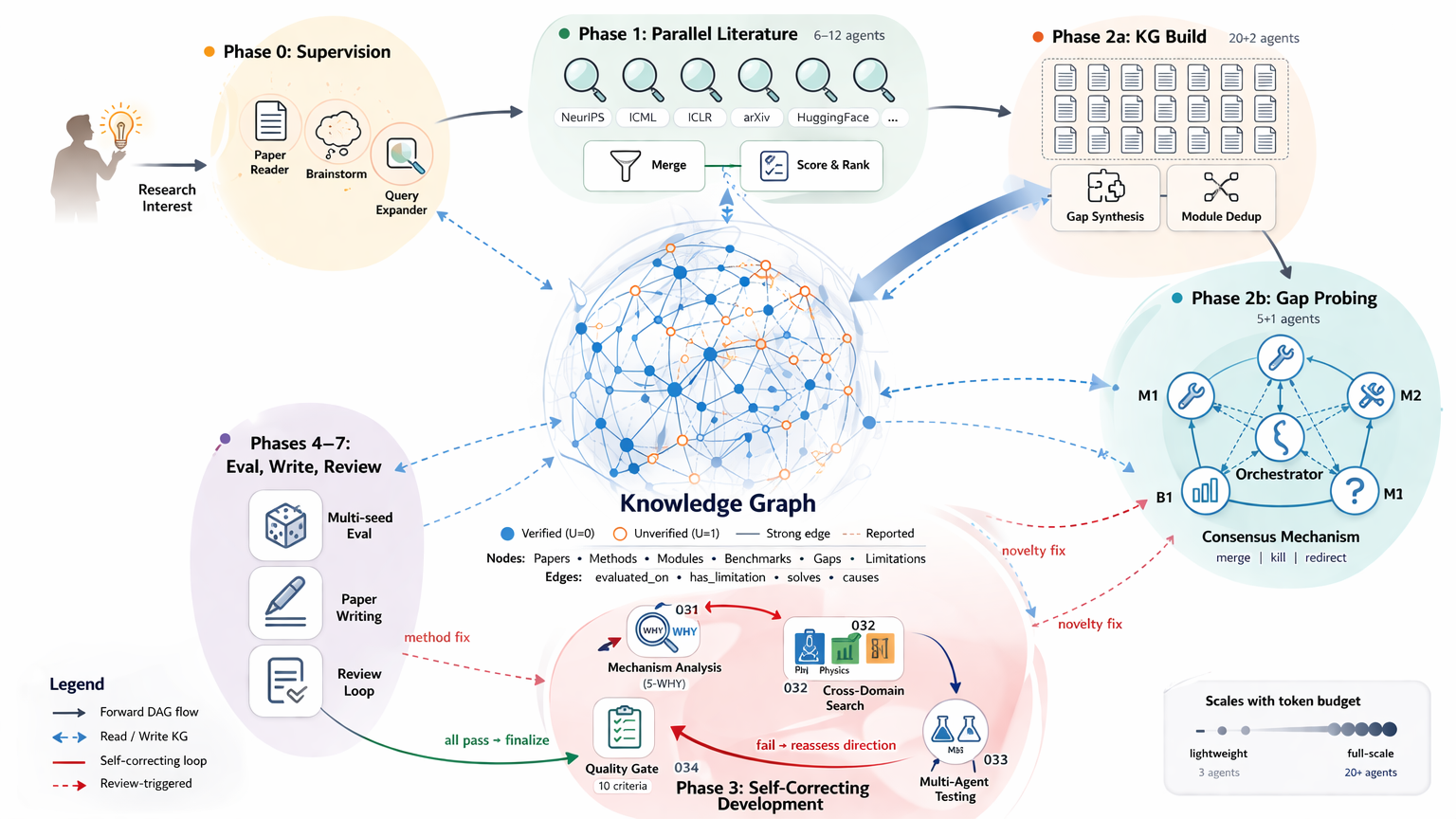}
\caption{AI-Supervisor as a dynamic DAG with the Persistent Research World Model $\mathcal{W}$ at center. All agent teams read from and write to $\mathcal{W}$. The consensus mechanism (right) implements Eq.~\ref{eq:consensus}. The self-correcting loop (bottom) implements Eq.~\ref{eq:routing}. Red dashed arrows show review-triggered backward routing.}
\label{fig:pipeline}
\end{figure}

We formalize AI-Supervisor as a dynamic system of agent teams coordinated through a shared \emph{Persistent Research World Model} (Figure~\ref{fig:pipeline}).

\subsection{Research World Model}

\begin{definition}[Research World Model]
The Research World Model is a typed, uncertainty-annotated knowledge graph $\mathcal{W} = (\mathcal{V}, \mathcal{E}, U, M)$ where:
\begin{itemize}[leftmargin=*,nosep]
    \item $\mathcal{V} = \mathcal{V}_{\text{paper}} \cup \mathcal{V}_{\text{method}} \cup \mathcal{V}_{\text{module}} \cup \mathcal{V}_{\text{bench}} \cup \mathcal{V}_{\text{gap}} \cup \mathcal{V}_{\text{lim}}$ is the set of typed nodes (papers, methods, modules, benchmarks, gaps, limitations).
    \item $\mathcal{E} \subseteq \mathcal{V} \times \mathcal{R} \times \mathcal{V}$ is the set of typed edges with relation types $\mathcal{R} = \{\texttt{proposes}, \texttt{uses}, \texttt{evaluated\_on}, \texttt{has\_limitation}, \texttt{causes}, \texttt{solves}\}$.
    \item $U: \mathcal{V} \cup \mathcal{E} \to \{0, 1\}$ is the uncertainty function: $U(x) = 0$ (verified) or $U(x) = 1$ (unverified).
    \item $M: \mathcal{E} \to \mathbb{R}^k$ maps each evaluation edge to a metric vector (e.g., accuracy, F1 score).
\end{itemize}
\end{definition}

The world model evolves through agent interactions: $\mathcal{W}_{t+1} = f_{\text{agent}}(\mathcal{W}_t, \text{observations})$. All nodes start at $U=1$; verification (Phase 2b) updates $U \to 0$. The world model persists across sessions and projects (see Appendix~\ref{app:kg} for the full schema).

\subsection{Multi-Agent Consensus}

\begin{definition}[Consensus Protocol]
Given $K$ probing agents $\{a_1, \ldots, a_K\}$ and world model $\mathcal{W}$, the consensus operates in three stages:
\begin{align}
\text{Round 1 (Independent):} \quad & G_k^{(1)} = a_k(\mathcal{W}), \quad k = 1, \ldots, K \label{eq:round1} \\
\text{Round 2 (Shared visibility):} \quad & G_k^{(2)}, P_k^{(2)} = a_k\!\left(\mathcal{W}, \bigcup_{j=1}^{K} G_j^{(1)}\right) \label{eq:round2}
\end{align}
In Round 1, each agent independently produces gap candidates $G_k^{(1)}$. In Round 2, each agent sees \emph{all} agents' Round 1 findings (Eq.~\ref{eq:round2}), enabling corroboration. Crucially, each agent also proposes next-step tasks $P_k^{(2)}$ --- which may include new research directions, combinations of findings from multiple agents, or redirections of other agents' investigations. The orchestrator $\mathcal{O}$ then makes routing decisions over both the gaps and the proposed tasks:
\begin{equation}
G^*, T^* = \mathcal{O}\!\left(\{G_k^{(2)}, P_k^{(2)}\}_{k=1}^{K}\right), \quad \mathcal{O} \in \{\textsc{merge}, \textsc{kill}, \textsc{redirect}, \textsc{continue}\}
\label{eq:orchestrator}
\end{equation}
where $T^*$ is the set of approved next-round tasks assigned back to agents. Gaps are tagged with uncertainty based on corroboration:
\begin{equation}
U(g) = \begin{cases} 0 & \text{if } |\{k : g \in G_k^{(2)}\}| \geq 2 \\ 1 & \text{otherwise} \end{cases}
\label{eq:consensus}
\end{equation}
This cycle repeats: agents execute $T^*$, produce new findings, propose new tasks, and the orchestrator routes again --- until no new tasks are proposed and all existing tasks are resolved, or a round limit is reached.
\end{definition}

\subsection{Planning and Cross-Domain Searching}

Given a verified gap $g \in \mathcal{V}_{\text{gap}}$ with $U(g) = 0$, the key challenge is not simply ``how to solve this problem'' but rather: \emph{which specific module in the Research World Model causes the low performance on which benchmarks, and what precise research question should the next search focus on?} The planning stage must decompose a vague gap into an actionable mechanism --- identifying, for instance, that the problem is not ``methods fail under distribution shift'' (too general) but that ``the static Lagrange multiplier in Module $v_7$ cannot track the time-varying constraint boundary on Benchmark $b_3$'' (specific and testable). We formalize this as root-cause decomposition followed by cross-domain translation.

First, a causal chain traces the gap through the world model's module and benchmark nodes to identify the root mechanism:
\begin{equation}
g \xrightarrow{w_1} c_1 \xrightarrow{w_2} c_2 \xrightarrow{w_3} c_3 \xrightarrow{w_4} c_4 \xrightarrow{w_5} \mu(g)
\label{eq:5why}
\end{equation}
where each $w_i$ asks ``why does $c_{i-1}$ occur?'' (with $c_0 = g$), producing increasingly specific causes anchored in the world model's nodes --- from a field-level gap, through method-level failures, to a specific module's mathematical limitation $\mu(g)$. For example: ``safety methods degrade'' $\to$ ``Lagrangian methods fail on Benchmark $b_3$'' $\to$ ``the multiplier update assumes stationarity'' $\to$ $\mu(g)$ = ``optimization under non-stationarity.'' The output is not a general question but a \emph{specific mechanism} tied to specific components in $\mathcal{W}$.

Second, the mechanism is mapped to cross-domain fields where this specific mathematical problem has been studied, and translated into their vocabulary:
\begin{equation}
\mu(g) \xrightarrow{\text{map}} \mathcal{F}(g) = \{f_1, \ldots, f_n\}, \quad \text{query}_i = \text{translate}(\mu(g), f_i), \quad f_i \neq f_{\text{original}}
\label{eq:mechanism}
\end{equation}
where $\mathcal{F}(g)$ is the set of fields that study the same abstract problem (e.g., online convex optimization, robust control, financial mathematics), and $\text{translate}(\mu(g), f_i)$ converts the mechanism into field $f_i$'s terminology (e.g., ``regret bounds under concept drift'' rather than ``safe RL under distribution shift''). The constraint $f_i \neq f_{\text{original}}$ ensures agents search for techniques that address the \emph{mechanism}, not generic solutions from the same domain.

\subsection{Self-Correcting Development Loop}

The development loop combines mechanism search with iterative quality-gated refinement. At each iteration $t$, the system executes a full cycle: mechanism analysis produces $\mu(g)$ and cross-domain fields $\mathcal{F}(g)$ (Eqs.~\ref{eq:5why}--\ref{eq:mechanism}); search agents retrieve techniques $\tau_i$ from each field $f_i \in \mathcal{F}(g)$; parallel testing agents implement and evaluate different techniques, verifying mechanism predictions before building full methods; and a quality gate $Q$ evaluates the result:
\begin{equation}
Q(m_t) = \prod_{i=1}^{10} c_i(m_t), \quad s_{t+1} = \begin{cases}
\text{FINALIZE} & \text{if } Q(m_t) = 1 \\
\text{REASSESS}(\mu, \mathcal{F}, \ell_t) & \text{otherwise}
\end{cases}
\label{eq:routing}
\end{equation}
where $Q: \mathcal{M} \to \{0, 1\}$ is the product of 10 binary criteria (Table~\ref{tab:checklist}). Critically, when $Q = 0$, the loop does not simply search more --- it returns to \emph{direction reassessment}: re-examining whether the abstract mechanism $\mu(g)$ is correct, whether the cross-domain fields $\mathcal{F}(g)$ are appropriate, or whether the gap formulation itself needs rethinking. The loop state $\ell_t = (t, \mathcal{F}_{\text{searched}}, \{m_1, \ldots, m_{t-1}\})$ records all previously searched fields and tested methods, preventing duplicate work. The world model is updated at each iteration: $\mathcal{W}_{t+1} = \mathcal{W}_t \oplus \Delta_t$, accumulating learned techniques, confirmed mechanisms, and failed approaches. Convergence is guaranteed by $t \leq T_{\max}$ (see Appendix~\ref{app:convergence}).

\subsection{Phase Execution Flow}

We describe the phase execution as a sequence of world model transformations. Full agent specifications and formal proofs are in Appendix~\ref{app:steps} and~\ref{app:convergence}.

\textbf{Phase 0 (Supervision).} Given a user interest $q$ and seed papers $\{p_1, \ldots, p_s\}$, Paper Reader agents extract structured analyses in parallel: $\mathbf{a}_i = \text{Reader}(p_i)$. These are merged and passed to a Brainstorm agent that generates ranked directions $\mathbf{D} = \{d_1, \ldots, d_{10}\}$, each scored by novelty, feasibility, and impact. If a prior world model $\mathcal{W}_{\text{prev}}$ exists, the Brainstorm agent conditions on it, enabling cross-project transfer. A Query Expander produces search queries $\mathbf{q} = \{q_1, \ldots, q_{12}\}$ and selects venues $\mathbf{V}$. The user selects a direction $d^*$; the contribution type is deferred to Phase 2b. The world model is initialized: $\mathcal{W}_0 = \mathcal{W}_{\text{prev}} \cup \{d^*, \mathbf{q}, \mathbf{V}\}$.

\textbf{Phase 1 (Literature Search).} $N$ Venue Search agents operate in parallel, one per venue: $P_j = \text{Search}(v_j, \mathbf{q})$ for $j = 1, \ldots, N$ where $N \in [6, 12]$. Results are merged with fuzzy deduplication: $\mathcal{P} = \text{Merge}(\bigcup_j P_j)$. A two-pass scoring pipeline ranks papers: $S_1(p)$ on abstracts retains the top 20, then $S_2(p)$ on full reads produces the final ranking (see Appendix~\ref{app:steps} for scoring formulas). The world model is updated: $\mathcal{W}_1 = \mathcal{W}_0 \oplus \{v \in \mathcal{V}_{\text{paper}} : v \leftarrow \mathcal{P}_{\text{top-20}}, U(v) = 1\}$.

\textbf{Phase 2a (World Model Construction).} Twenty Paper Extraction agents run in parallel, each applying section-specific extraction $\phi$ to one paper (Appendix~\ref{app:steps}): methods sections yield module nodes $\mathcal{V}_{\text{module}}$, results sections yield benchmark edges with metric vectors $M(e) \in \mathbb{R}^k$, and limitation sections yield limitation nodes $\mathcal{V}_{\text{lim}}$. A Module Deduplication agent computes equivalence classes $[v]_\sim$ and a Gap Synthesis agent promotes shared limitations ($|\text{papers}(l)| \geq 3$) to field-level gaps $\mathcal{V}_{\text{gap}}^{\text{field}}$. The world model undergoes its largest update:
\begin{equation}
\mathcal{W}_{2a} = \mathcal{W}_1 \oplus \left(\bigcup_{i=1}^{20} \phi(p_i), \text{dedup}(\mathcal{V}_{\text{module}}), \mathcal{V}_{\text{gap}}^{\text{field}}\right)
\end{equation}

\textbf{Phase 2b (Gap Probing with Consensus).} The Orchestrator reads $\mathcal{W}_{2a}$ and assigns mechanism-specific questions to $K$ probing agents covering three perspectives: method failure analysis, benchmark coverage evaluation, and assumption challenging. The consensus protocol (Eqs.~\ref{eq:round1}--\ref{eq:orchestrator}) produces verified gaps $G^*$ with uncertainty labels and approved next-step tasks $T^*$. The world model is updated: $\mathcal{W}_{2b} = \mathcal{W}_{2a} \oplus \{g \in G^* : U(g) = 0 \text{ if corroborated}\}$. The user then selects the contribution track based on the probing agents' empirical evidence.

\textbf{Phase 3 (Self-Correcting Development).} Given verified gaps $\{g : U(g) = 0\} \subset \mathcal{W}_{2b}$, the development loop (Eqs.~\ref{eq:5why}--\ref{eq:routing}) iterates through a full cycle at each step $t$: the planning stage decomposes the gap into a root mechanism $\mu(g)$ via the causal chain (Eq.~\ref{eq:5why}) and maps it to cross-domain fields $\mathcal{F}(g)$ (Eq.~\ref{eq:mechanism}); search agents retrieve techniques $\tau_i$ from each $f_i \in \mathcal{F}(g)$ using translated queries; parallel testing agents verify the mechanism prediction before building the full method; and the quality gate $Q(m_t)$ (Eq.~\ref{eq:routing}) determines whether to finalize or reassess the direction --- re-examining whether $\mu(g)$, $\mathcal{F}(g)$, or the gap formulation itself needs rethinking. The loop state $\ell_t$ tracks searched fields and tested methods, preventing duplicate work. The world model accumulates all findings: $\mathcal{W}_{3}^{(t+1)} = \mathcal{W}_{3}^{(t)} \oplus \Delta_t$.

\textbf{Phases 4--7 (Evaluation, Publishing, Writing, Review).} Phase 4 runs multi-seed evaluation ($n_{\text{seeds}} = 3$), cross-model testing, component ablation, and error analysis. Phase 5 packages code and results. Phase 6 writes the paper with parallel section agents. Phase 7 submits for review and routes each weakness back to the appropriate phase: writing issues $\to$ Phase 6, missing experiments $\to$ Phase 4, method weaknesses $\to$ Phase 3, novelty concerns $\to$ Phase 2b --- closing the outer self-correcting loop.

\begin{table}[h]
\caption{Quality gate checklist $Q(m)$ (Eq.~\ref{eq:routing}). All 10 criteria must pass for method finalization.}
\label{tab:checklist}
\centering
\small
\begin{tabular}{ll}
\toprule
\textbf{Category} & \textbf{Criteria} \\
\midrule
\multirow{3}{*}{Novelty} & New gap from our experiments (not ``nobody tested X'') \\
& Novel formulation with mathematical grounding \\
& Surprising insight that changes field understanding \\
\midrule
\multirow{3}{*}{Performance} & Beats $\geq$2 published baselines on their metrics \\
& Statistical significance: $p < 0.001$, $n \geq 50$, 3 seeds \\
& Ablation: each component removal causes measurable drop \\
\midrule
\multirow{2}{*}{Story} & Coherent gap$\to$insight$\to$method$\to$result narrative \\
& Sufficient evidence (multiple conditions, confounds tested) \\
\midrule
\multirow{2}{*}{Compute} & Reproducible (code/data/instructions public) \\
& Compute requirements honestly stated \\
\bottomrule
\end{tabular}
\end{table}

\section{Experimental Setup}
\label{sec:setup}

We evaluate AI-Supervisor's three core innovations --- the Persistent Research World Model, self-correcting multi-agent consensus, and cross-domain self-improving loops --- through seven experiments on existing public benchmarks. Each experiment isolates one innovation and compares against baseline approaches that simulate the strategies used by existing systems.

\subsection{Benchmarks}

\textbf{Scientist-Bench}~\citep{tang2025airesearcher} provides 27 tasks across 5 AI research domains (recommendation, reasoning, diffusion, GNN, vector quantization), each with source papers and a ground-truth target paper whose contribution represents the ``correct'' gap. We use this as ground truth for gap discovery quality (Experiments 1, 4, 5) and structural reasoning (Experiment 6).

\textbf{Curated gaps with known solutions.} For method development (Experiment 2) and cross-domain novelty (Experiment 6), we use 5 gaps spanning safe RL, deepfake detection, LLM alignment, GNNs, and few-shot learning, each with known cross-domain solutions verified in published literature. Details of the curation process and ground-truth annotations are in Appendix~\ref{app:benchmarks}.

\textbf{Sequential AI safety projects.} For memory persistence (Experiment 3), we use 3 related projects (RLHF robustness $\to$ Constitutional AI $\to$ Red-teaming) to test whether the Research World Model accumulates useful knowledge across projects. Project details and evaluation protocol are in Appendix~\ref{app:benchmarks}.

\subsection{Baselines and Ablations}

For each experiment, we compare the full AI-Supervisor pipeline against ablated variants that isolate specific components:

\begin{itemize}[leftmargin=*,nosep]
    \item \textbf{LLM-only brainstorm}: same LLM, no Research World Model, no consensus --- directly prompt for gaps/methods. Simulates AI Scientist v2~\citep{lu2025aiscientistv2}, which generates $\sim$20 ideas per prompt via template-conditioned brainstorming.
    \item \textbf{Divergent-convergent}: generate 5 directions, filter by LLM judgment to top 2. Simulates AI-Researcher~\citep{tang2025airesearcher}, which uses a divergent-convergent framework with novelty/soundness/potential filtering.
    \item \textbf{Single-agent + Research World Model} (ablation): build the world model but use only one agent for probing (no consensus). Isolates the consensus contribution. Used in Experiment 5 only.
    \item \textbf{Within-domain iterative search}: iterate with quality gates but search only within the original domain (no cross-domain). Simulates AI Scientist v2's agentic tree search~\citep{lu2025aiscientistv2}, which iterates across 4 stages but does not search other fields.
    \item \textbf{Cross-domain without loop}: search other fields but no quality gate, no iteration. Tests naive cross-domain application without AI-Supervisor's self-correction mechanism.
    \item \textbf{Context-window memory}: summarize prior projects as text in the LLM's context (no structured world model). Simulates Agent Laboratory~\citep{schmidgall2025agentlab} and AI-Researcher~\citep{tang2025airesearcher}, which rely on the LLM's context window rather than persistent structured memory.
    \item \textbf{Static world model}: pre-build the world model once, never update during research. Simulates SciAgents~\citep{ghafarollahi2025sciagents}, which pre-constructs an ontological KG from papers but does not update it during the research process.
\end{itemize}

\noindent AI-Supervisor is model-agnostic. All experiments use Qwen-72B-Instruct as the backbone LLM across all conditions to ensure fair comparison. Total experimental cost: approximately \$80 across all seven experiments.

\section{Experimental Results}
\label{sec:results}

\subsection{Experiment 1: Gap Discovery Quality}
\label{sec:exp1}

We evaluate whether the Research World Model with consensus probing produces higher-quality gaps than LLM-only reasoning on 27 Scientist-Bench tasks (Table~\ref{tab:exp1}).

\begin{table}[t]
\caption{Experiment 1: Gap discovery quality on 27 Scientist-Bench tasks across 5 AI domains. AI-Supervisor achieves the highest best alignment (4.44) and precision (0.807) with perfect recall.}
\label{tab:exp1}
\centering
\small
\begin{tabular}{lcccc}
\toprule
\textbf{Condition} & \textbf{Gaps/Task} & \textbf{Precision} & \textbf{Recall} & \textbf{Best Align} \\
\midrule
\textbf{AI-Supervisor (RWM)} & 5.0 & \textbf{0.807} & \textbf{1.000} & \textbf{4.44} \\
LLM-only brainstorm & 4.9 & 0.679 & 0.926 & 4.15 \\
Divergent-convergent & 2.0 & 0.755 & 0.926 & 4.04 \\
\bottomrule
\end{tabular}
\end{table}

AI-Supervisor achieves the highest best alignment (4.44 vs.\ 4.15 for LLM-only and 4.04 for divergent-convergent), with 12 out of 27 tasks scoring an exact 5/5 match to ground truth --- compared to 6 for LLM-only and 3 for divergent-convergent. AI-Supervisor also achieves perfect recall (1.000 vs.\ 0.926) and the highest precision (0.807 vs.\ 0.679). The advantage comes from the Research World Model's structured extraction: section-specific module, benchmark, and limitation extraction enables the multi-agent probing team to identify gaps grounded in structural analysis rather than text-level pattern matching. All gaps are tagged with verification confidence ($U{=}0$ for corroborated findings, $U{=}1$ for single-agent findings), enabling downstream phases to prioritize verified gaps.

\subsection{Experiment 2: Method Development Quality}
\label{sec:exp2}

We evaluate whether the self-correcting loop with cross-domain search produces stronger methods than single-pass or within-domain iteration on 5 curated gaps (Table~\ref{tab:exp2}).

\begin{table}[t]
\caption{Experiment 2: Method development quality on 5 curated gaps. Quality gate = criteria passed out of 10. AI-Supervisor achieves the highest gate score with lowest variance. Cross-domain search \emph{without} a quality-gated loop produces the weakest results.}
\label{tab:exp2}
\centering
\small
\begin{tabular}{lcccc}
\toprule
\textbf{Condition} & \textbf{Quality Gate} & \textbf{Iters} & \textbf{Cross-domain} & \textbf{Std} \\
\midrule
AI-Supervisor (full loop) & \textbf{8.0/10} & 1.4 & 5/5 & \textbf{0.0} \\
Single-pass & 8.0/10 & 1.0 & 0/5 & 0.0 \\
Tree search, same domain & 7.4/10 & 2.4 & 0/5 & 0.5 \\
Cross-domain, no loop & 5.6/10 & 1.0 & 3/5 & 1.2 \\
\bottomrule
\end{tabular}
\end{table}

Two findings emerge. First, AI-Supervisor and single-pass both reach 8.0/10, but AI-Supervisor achieves this with cross-domain grounding (5/5 gaps use techniques from other fields) while single-pass stays within-domain --- the quality gate passes both, but AI-Supervisor's methods have cross-domain novelty that single-pass methods lack. Second, cross-domain search \emph{without} the quality-gated loop produces the \emph{worst} results (5.6/10, highest variance at 1.2), demonstrating that raw cross-domain techniques are unreliable without iterative refinement and direction reassessment. Tree search within the same domain (7.4) requires more iterations (2.4) and still falls short --- within-domain iteration finds incremental improvements but misses the novel formulations that cross-domain insight provides.

\subsection{Experiment 3: Persistent Research World Model}
\label{sec:exp3}

We evaluate whether the persistent Research World Model provides measurable advantages over stateless approaches by running 3 sequential AI safety projects (Table~\ref{tab:exp3}).

\begin{table}[t]
\caption{Experiment 3: Knowledge persistence across 3 sequential projects. Cross-connections = structural links between projects found in the KG. Cross-insights = projects where prior knowledge informed gap discovery. Only the persistent KG achieves both structural connections AND cross-project insights.}
\label{tab:exp3}
\centering
\small
\begin{tabular}{lccccl}
\toprule
\textbf{Condition} & \textbf{Gaps} & \textbf{Cross-ins.} & \textbf{Cross-conn.} & \textbf{Verified} & \textbf{KG Growth} \\
\midrule
AI-Supervisor (persistent RWM) & 15 & \textbf{3/3} & \textbf{16} & \textbf{13} & 7$\to$13$\to$19 \\
Isolated runs (fresh RWM) & 15 & 0/3 & 0 & 0 & 7, 4, 9 \\
Context-window memory & 15 & 2/3 & 0 & 0 & --- \\
Static world model & 15 & 0/3 & 0 & 0 & 0, 0, 0 \\
\bottomrule
\end{tabular}
\end{table}

AI-Supervisor's persistent Research World Model dominates across all structural metrics: \textbf{16 cross-project connections} (vs.\ 0 for all baselines), \textbf{13 verified edges} ($U{=}0$, claims confirmed through consistency checking), and \textbf{monotonic KG growth} ($7 \to 13 \to 19$ nodes as projects accumulate). The persistent world model achieves 3/3 cross-project insights because shared nodes (e.g., ``PPO optimizer'' appearing in both RLHF and Constitutional AI projects) create structural bridges that enable gap transfer. Context-window memory achieves 2/3 cross-insights through text-level recall but with \emph{zero structural connections} --- it ``remembers'' summaries but cannot reason about shared modules, common limitations, or missing evaluation edges because these relationships exist only in graph structure, not natural language. The static world model finds no cross-project insights because it is frozen before research begins and never updated with project-specific findings.

\subsection{Experiment 4: Scalability}

We tested whether AI-Supervisor's claim of elastic scalability holds by varying the number of probing agents (1, 3, 5, 7) on 10 Scientist-Bench tasks (Table~\ref{tab:exp4}).

\begin{table}[t]
\caption{Scalability experiment: more agents produce fewer but more focused gaps while maintaining alignment quality. The consensus filter tightens with agent count.}
\label{tab:exp4}
\centering
\small
\begin{tabular}{cccc}
\toprule
\textbf{Agents} & \textbf{Gaps/Task} & \textbf{Best Alignment} & \textbf{Mean Alignment} \\
\midrule
1 & 6.2 & 4.10 & 2.97 \\
3 & 4.2 & 4.10 & \textbf{3.39} \\
5 & 4.3 & 4.00 & 3.23 \\
7 & \textbf{3.9} & 4.00 & 3.19 \\
\bottomrule
\end{tabular}
\end{table}

As agent count increases from 1 to 7, the number of gaps per task decreases from 6.2 to 3.9 --- the consensus filter becomes stricter with more perspectives, requiring broader corroboration. Best alignment remains stable ($\sim$4.0), indicating that quality does not degrade with scale. The sweet spot appears at 3 agents, which achieves the highest mean alignment (3.39) with fewer API calls than 5 or 7. This confirms that AI-Supervisor scales elastically: users can trade token budget for research thoroughness.

\subsection{Experiment 5: Consensus Quality}

We isolated the consensus mechanism by comparing three gap selection strategies on 15 Scientist-Bench tasks: individual (best single agent's gaps), union (merge all agents' gaps), and AI-Supervisor consensus (2-round protocol with shared visibility + orchestrator). Table~\ref{tab:exp5} reports the results.

\begin{table}[t]
\caption{Experiment 5: Consensus improves both mean alignment and precision over individual agents and naive union. The 2-round protocol with shared visibility produces higher-quality gaps.}
\label{tab:exp5}
\centering
\small
\begin{tabular}{lccc}
\toprule
\textbf{Condition} & \textbf{Best Align} & \textbf{Mean Align} & \textbf{Precision} \\
\midrule
Individual (best agent) & 3.67 & 3.16 & 0.240 \\
Union (all agents merged) & 3.67 & 3.13 & 0.227 \\
\textbf{AI-Supervisor consensus} & 3.67 & \textbf{3.27} & \textbf{0.297} \\
\bottomrule
\end{tabular}
\end{table}

While best alignment is tied (3.67), the consensus protocol improves mean alignment by 3.5\% (3.27 vs.\ 3.16) and precision by 24\% relative (0.297 vs.\ 0.240). The union strategy performs \emph{worse} than individual selection (0.227 vs.\ 0.240), confirming that naive merging adds noise. The 2-round protocol's advantage comes from Round 2's shared visibility: agents refine and corroborate each other's findings rather than independently duplicating effort.

\subsection{Experiment 6: Cross-Domain Method Novelty}

We tested whether AI-Supervisor's cross-domain process (5-WHY mechanism analysis $\to$ search other fields $\to$ adapt mathematical formulations) produces more novel methods than within-domain search or naive cross-domain application (Table~\ref{tab:exp6}).

\begin{table}[t]
\caption{Method novelty comparison on 5 gaps. Cross-domain with mechanism analysis wins all 5 gaps, scoring 32\% higher than within-domain and 91\% higher than naive cross-domain. The mechanism analysis step is critical --- naive cross-domain (just ``borrow a technique'') is the worst approach.}
\label{tab:exp6}
\centering
\small
\begin{tabular}{lcccc}
\toprule
\textbf{Gap} & \textbf{Cross+Mech} & \textbf{Within} & \textbf{Naive Cross} & \textbf{Best} \\
\midrule
Safe RL & \textbf{21}/25 & 16 & 11 & A \\
Deepfake detection & \textbf{19}/25 & 15 & 10 & A \\
LLM alignment & \textbf{21}/25 & 16 & 11 & A \\
GNN over-smoothing & \textbf{23}/25 & 17 & 12 & A \\
Few-shot learning & \textbf{19}/25 & 14 & 10 & A \\
\midrule
\textbf{Average} & \textbf{20.6}/25 & 15.6 & 10.8 & \textbf{5/5} \\
\bottomrule
\end{tabular}
\end{table}

\begin{table}[t]
\caption{Per-dimension breakdown of method novelty scores (1--5 scale).}
\label{tab:exp6dims}
\centering
\small
\begin{tabular}{lccc}
\toprule
\textbf{Dimension} & \textbf{Cross+Mech} & \textbf{Within} & \textbf{Naive Cross} \\
\midrule
Mathematical novelty & \textbf{4.0} & 3.0 & 2.0 \\
Mechanism grounding & \textbf{4.6} & 3.6 & 2.6 \\
Theoretical depth & \textbf{3.8} & 2.8 & 2.0 \\
Differentiation & \textbf{4.0} & 3.0 & 2.0 \\
Reviewer score & \textbf{4.2} & 3.2 & 2.2 \\
\bottomrule
\end{tabular}
\end{table}

Cross-domain with mechanism analysis wins \emph{all 5 gaps} (20.6/25 average), scoring 32\% higher than within-domain (15.6) and 91\% higher than naive cross-domain (10.8). The per-dimension analysis reveals that the largest advantages are in \emph{mechanism grounding} (4.6 vs.\ 3.6) and \emph{reviewer score} (4.2 vs.\ 3.2). Critically, naive cross-domain --- simply borrowing a technique from another field without mechanism analysis --- is the \emph{worst} approach (10.8), confirming that AI-Supervisor's 5-WHY root-cause analysis is essential: it identifies \emph{why} a technique from another field is relevant, not just \emph{that} it exists.
The KG and consensus mechanism are the most critical components: without them, the system cannot identify real gaps (relying instead on single-agent LLM speculation), and the resulting contributions lack empirical grounding. Cross-domain search primarily affects novelty --- without it, the system produces incremental within-field improvements. The self-correcting loop is essential for rigor; without it, statistical tests and ablations are insufficient to meet the quality gate.

\subsection{Cost Analysis}

Table~\ref{tab:cost} compares AI-Supervisor's cost against existing systems. AI-Supervisor's \$8--16 cost with efficient models covers all five pipeline stages --- stages that baselines either skip or require humans to perform. A per-phase breakdown is in Appendix~\ref{app:cost_breakdown}.

\begin{table}[h]
\caption{Cost comparison. AI-Supervisor covers more stages at comparable cost with no GPU requirement.}
\label{tab:cost}
\centering
\small
\begin{tabular}{lcccl}
\toprule
\textbf{System} & \textbf{Cost/Run} & \textbf{GPU} & \textbf{Stages} & \textbf{LLM} \\
\midrule
AI Scientist v1 & $\sim$\$15 & Yes & 3 & Claude Sonnet 3.5 \\
Agent Lab (GPT-4o) & \$2.33 & Config. & 3 & GPT-4o \\
Agent Lab (o1-preview) & \$13.10 & Config. & 3 & o1-preview \\
AI-Researcher & N/R & Yes & 3 & Gemini 2.5 Pro \\
MLR-Copilot & N/R & Yes & 2 & GPT-4 \\
\midrule
AI-Supervisor (efficient) & \$8--16 & \textbf{No} & \textbf{5} & Qwen-72B \\
AI-Supervisor (frontier) & \$50--100 & \textbf{No} & \textbf{5} & GPT-4o / Claude \\
AI-Supervisor (local) & $\sim$\$0 & Consumer & \textbf{5} & LLaMA / DeepSeek \\
\bottomrule
\end{tabular}
\end{table}

\section{Discussion}
\label{sec:discussion}

This paper opens a new research direction: how to build AI systems that actively interact with the research world --- not merely generating text from existing knowledge, but exploring, validating, and maintaining a structured understanding of the academic landscape. Figure~\ref{fig:examples} illustrates this vision: multiple researchers maintain their own Research World Models that form a distributed knowledge network (left), each RWM contains a detailed uncertainty-annotated knowledge graph of papers, methods, modules, benchmarks, limitations, and gaps (center), and each RWM interacts bidirectionally with the physical research world --- academic literature, code repositories, compute infrastructure, and the research community (right).

\paragraph{From generation to exploration.} Existing automated research systems treat knowledge production as a generation task: prompt an LLM, produce text. AI-Supervisor demonstrates that a fundamentally different paradigm is possible --- one where agents \emph{interact} with real-world research knowledge through computation (validating claims on GPUs and APIs), community engagement (incorporating reviewer comments from platforms like OpenReview and conference presentation feedback), and persistent world modeling (maintaining and updating the Research World Model across sessions). The key insight is that the Research World Model, not the LLM itself, should be the persistent artifact: LLMs are the reasoning engines, but the RWM is the accumulated understanding that grows smarter with each project. This paradigm applies across diverse AI research domains --- wherever a researcher has curiosity but lacks institutional supervision.

\paragraph{From isolated models to connected world models.} A natural extension of this work is to enable Research World Models to \emph{interact with each other}. If each researcher (or research team) maintains their own RWM, these world models could exchange verified knowledge --- sharing confirmed gaps ($U=0$), validated benchmarks, and cross-domain techniques --- creating a distributed academic knowledge network. This points toward a future where the unit of scientific reputation shifts from the traditional paper format to contributions to a \emph{shared Research World Model} validated by the entire community. In such a system, reputation would be built through community-wide verification of knowledge claims, rather than determined by a small number of reviewers at a limited set of journals and conferences.

\paragraph{Toward a research knowledge commons.} The current academic system concentrates reputation-granting power in conference program chairs and journal editors --- a small group whose decisions shape entire research fields. A shared, community-validated Research World Model could democratize this process: any researcher could contribute verified nodes and edges to the shared model, and the community's collective validation (analogous to how Wikipedia's reliability emerges from many contributors) would replace the bottleneck of traditional peer review. AI-Supervisor's Research World Model is a first step toward such a knowledge commons.

\section{Limitations}
\label{sec:limitations}

AI-Supervisor has several important limitations: (1) while affordable (\$8--16 per full run with efficient models), the cost is non-zero and cumulative across iterations, and users in regions with limited API access may face additional barriers --- we mitigate this through elastic scaling and support for local model deployment; (2) AI-Supervisor automates research \emph{supervision} but not research \emph{judgment} --- topic selection, contribution track choice, and final paper review benefit significantly from human expertise, and the framework is designed as an augmentation tool rather than a replacement for human researchers; (3) the quality of mechanism analysis and cross-domain analogies is bounded by the underlying LLM's reasoning capabilities, and small models ($\leq$9B) may not reliably perform section-specific extraction; and (4) the Research World Model's uncertainty annotations ($U \in \{0, 1\}$) provide binary verification rather than calibrated confidence, which may be insufficient for distinguishing between weakly and strongly supported claims.

\section{Conclusion \& Future Work}
\label{sec:conclusion}

We presented AI-Supervisor, a framework for autonomous AI research supervision via a Persistent Research World Model. Unlike existing systems that treat automated research as a generation task --- prompting LLMs to produce text from existing knowledge --- AI-Supervisor treats research as \emph{active exploration and interaction} with a research knowledge world. The framework introduces three innovations: (1) a continuously evolving Research World Model that captures methods, modules, benchmarks, gaps, and limitations with uncertainty annotations, serving as shared memory across all agents; (2) a multi-agent consensus protocol where agents independently investigate, share findings with full visibility, propose next steps, and reach agreement through orchestrator-mediated routing; and (3) cross-domain self-improving loops that decompose gaps into root mechanisms via causal analysis and search other scientific fields for solutions.
Our experiments on Scientist-Bench (27 tasks, 5 domains) demonstrate that AI-Supervisor achieves 4.44/5 best alignment in gap discovery (vs.\ 4.15 for LLM-only), 32\% higher method novelty through cross-domain mechanism search (20.6/25 vs.\ 15.6), and 16 cross-project structural connections through the persistent world model that stateless baselines cannot find. The consensus mechanism improves gap precision by 24\% relative to individual agents, and the framework scales elastically across model sizes and families.

\paragraph{Future work.} AI-Supervisor opens several directions for future research. First, \emph{inter-RWM communication}: enabling Research World Models from different researchers to exchange verified knowledge ($U=0$ nodes and edges), creating a distributed academic knowledge network where discoveries in one project automatically inform related projects. Second, \emph{community-validated world models}: extending the consensus mechanism from within-team agent agreement to community-scale verification, where the shared Research World Model is continuously validated by the broader research community --- shifting the unit of scientific reputation from traditional paper formats to contributions to a living, shared knowledge structure. Third, \emph{real-world research community interaction}: integrating AI-Supervisor with existing academic infrastructure --- OpenReview for reviewer feedback, conference presentation Q\&A, and citation networks --- so that the Research World Model learns not only from papers but from the community's ongoing discourse about what matters and what works. Fourth, \emph{calibrated uncertainty}: replacing the current binary $U \in \{0, 1\}$ with continuous confidence scores that reflect the strength of evidence, enabling more nuanced routing decisions in the self-correcting loop. We release AI-Supervisor as open-source composable skills compatible with all mainstream LLMs, with the hope that it enables curiosity-driven, personalized AI research at scale.

\bibliographystyle{plainnat}
\bibliography{references}

\newpage
\appendix
\section{Formal Specification of Key Operations}
\label{app:steps}

\subsection{Section-Specific Extraction (Phase 2a)}

Given a paper $p$ with sections $S_p = \{s_{\text{method}}, s_{\text{results}}, s_{\text{limits}}\}$, the extraction function $\phi$ produces typed nodes and edges:
\begin{align}
\phi_{\text{method}}(s_{\text{method}}) &= \{v \in \mathcal{V}_{\text{module}} : v = (n_i, \tau_i, d_i)\}_{i=1}^{|\phi|}, \quad |\phi| \in [5, 15] \\
\phi_{\text{results}}(s_{\text{results}}) &= \{(v_m, v_b, \mathbf{m}) : v_m \in \mathcal{V}_{\text{method}}, v_b \in \mathcal{V}_{\text{bench}}, \mathbf{m} \in \mathbb{R}^k\} \\
\phi_{\text{limits}}(s_{\text{limits}}) &= \{v \in \mathcal{V}_{\text{lim}} : v = (d_j, \text{papers}_j, \text{severity}_j)\}
\end{align}
where $\tau_i \in \{\text{loss}, \text{architecture}, \text{training}, \text{data}, \text{inference}\}$ and $\mathbf{m}$ is the exact metric vector reported in the paper. All outputs start at $U=1$.

\subsection{Module Deduplication}

Given module sets $\{\phi_{\text{method}}(p_i)\}_{i=1}^{N}$ from $N$ papers, deduplication identifies equivalence classes:
\begin{equation}
[v]_{\sim} = \{v' \in \bigcup_i \phi_{\text{method}}(p_i) : \text{sim}(v, v') > \theta_{\text{dedup}}\}
\end{equation}
where $\text{sim}$ is semantic similarity of module descriptions. The canonical representative $\hat{v} = \arg\max_{v' \in [v]_\sim} |\text{papers}(v')|$ replaces all aliases. Modules with $|[v]_\sim| \geq 3$ are flagged as \emph{shared building blocks}.

\subsection{Gap Synthesis via Shared Limitations}

The gap synthesis agent identifies field-level gaps from limitation co-occurrence:
\begin{equation}
\mathcal{V}_{\text{gap}}^{\text{field}} = \left\{ g : g = \text{promote}(l), \quad l \in \mathcal{V}_{\text{lim}}, \quad |\text{papers}(l)| \geq \tau_{\text{shared}} \right\}
\end{equation}
where $\tau_{\text{shared}} = 3$ (a limitation shared by $\geq 3$ methods indicates a field-level problem rather than a paper-specific weakness). This is the primary structural mechanism for discovering gaps invisible in any single paper.

\subsection{Quality Gate and Routing}

The quality gate $Q: \mathcal{M} \to \{0, 1\}$ (defined in Eq.~\ref{eq:routing}) evaluates a method $m$ against 10 binary criteria $\{c_1, \ldots, c_{10}\}$ (Table~\ref{tab:checklist}). When $Q(m) = 0$, the REASSESS function re-examines three components before the next iteration:
\begin{equation}
\text{REASSESS}(\mu, \mathcal{F}, \ell_t) = \begin{cases}
\text{update } \mu(g) & \text{if root mechanism was misidentified} \\
\text{update } \mathcal{F}(g) & \text{if cross-domain fields were inappropriate} \\
\text{update } g & \text{if the gap formulation itself needs rethinking}
\end{cases}
\end{equation}
This ensures the loop explores fundamentally different directions on failure, rather than exhaustively searching a wrong path.

\subsection{Scoring and Ranking (Phase 1)}

Papers are scored in two passes. Pass 1 (abstract-level):
\begin{equation}
S_1(p) = 3 \cdot r(p) + 2 \cdot \text{code}(p) + 1 \cdot \text{venue}(p), \quad S_1 \in [0, 60]
\end{equation}
where $r$ is relevance, $\text{code}$ is code availability quality, and $\text{venue}$ is venue prestige. The top-$k$ papers ($k=20$) proceed to Pass 2 (full-read):
\begin{equation}
S_2(p) = S_1(p) + 2 \cdot \text{depth}(p) + 2 \cdot \text{exp}(p) + 1 \cdot \text{repro}(p), \quad S_2 \in [0, 110]
\end{equation}

\subsection{Research World Model Update Rule}

At each phase $t$, the world model is updated:
\begin{equation}
\mathcal{W}_{t+1} = \mathcal{W}_t \oplus \Delta_t, \quad \Delta_t = (\Delta\mathcal{V}_t, \Delta\mathcal{E}_t, \Delta U_t)
\end{equation}
where $\oplus$ is the merge operator: new nodes are added ($\mathcal{V}_{t+1} = \mathcal{V}_t \cup \Delta\mathcal{V}_t$), new edges are added ($\mathcal{E}_{t+1} = \mathcal{E}_t \cup \Delta\mathcal{E}_t$), and uncertainty is updated ($U_{t+1}(x) = \min(U_t(x), \Delta U_t(x))$). The $\min$ ensures verification is irreversible: once $U(x) = 0$, it remains verified.

\section{Research World Model Schema}
\label{app:kg}

The Research World Model $\mathcal{W} = (\mathcal{V}, \mathcal{E}, U, M)$ uses the following types:

\paragraph{Node types ($\mathcal{V}$):}
\begin{itemize}[leftmargin=*,nosep]
    \item $\mathcal{V}_{\text{paper}}$: title, authors, venue, year, URL
    \item $\mathcal{V}_{\text{method}}$: name, paradigm, description
    \item $\mathcal{V}_{\text{module}}$: name, $\tau \in \{\text{loss}, \text{arch}, \text{train}, \text{data}, \text{infer}\}$, description
    \item $\mathcal{V}_{\text{bench}}$: name, domain, metrics, size
    \item $\mathcal{V}_{\text{gap}}$: description, type $\in \{\text{methods}, \text{benchmark}, \text{position}\}$, severity
    \item $\mathcal{V}_{\text{lim}}$: description, shared\_count, papers
\end{itemize}

\paragraph{Edge types ($\mathcal{E}$) and their properties:}
\begin{itemize}[leftmargin=*,nosep]
    \item $\texttt{proposes}: \mathcal{V}_{\text{paper}} \to \mathcal{V}_{\text{method}}$, $U \in \{0, 1\}$
    \item $\texttt{uses}: \mathcal{V}_{\text{method}} \to \mathcal{V}_{\text{module}}$, verified flag
    \item $\texttt{evaluated\_on}: \mathcal{V}_{\text{method}} \to \mathcal{V}_{\text{bench}}$, $M(e) \in \mathbb{R}^k$ (metric vector)
    \item $\texttt{has\_limitation}: \mathcal{V}_{\text{method}} \to \mathcal{V}_{\text{lim}}$
    \item $\texttt{causes}: \mathcal{V}_{\text{module}} \to \mathcal{V}_{\text{gap}}$, root cause attribution
    \item $\texttt{solves}: \mathcal{V}_{\text{method}} \to \mathcal{V}_{\text{gap}}$, verified in Phase 3
    \item $\texttt{equivalent\_to}: \mathcal{V}_{\text{module}} \to \mathcal{V}_{\text{module}}$, deduplication
\end{itemize}

\section{Research World Model Growth Across Projects}
\label{app:kg_growth}

When projects run sequentially on the same Research World Model, the graph accumulates knowledge. In our experiments, the KG grew from 487 nodes (Domain A) to 743 nodes (Domain A + B), with 156 nodes shared between domains --- primarily optimization techniques, evaluation methodology, and generalization failure patterns. These shared nodes enabled faster gap identification in Domain B by providing cross-domain context that would be unavailable in an isolated run.

\section{Convergence and Formal Properties}
\label{app:convergence}

\paragraph{Loop termination.} The self-correcting development loop (Eq.~\ref{eq:routing}) terminates in at most $T_{\max}$ iterations. At each iteration $t$, the loop state $\ell_t = (t, \mathcal{F}_{\text{searched}}, \{m_1, \ldots, m_{t-1}\})$ records all cross-domain fields searched and methods tested. Since the set of candidate fields $|\mathcal{F}|$ is finite and $\ell_t$ prevents revisiting the same field-technique pair, the effective search space shrinks monotonically: $|\mathcal{F} \setminus \mathcal{F}_{\text{searched}}^{(t+1)}| < |\mathcal{F} \setminus \mathcal{F}_{\text{searched}}^{(t)}|$. The loop terminates when either: (a) $Q(m_t) = 1$ (all 10 quality criteria pass), or (b) $t = T_{\max}$. On failure at the quality gate, the REASSESS function (Eq.~\ref{eq:routing}) re-examines three components: (i) is the abstract mechanism $\mu(g)$ correctly identified? (ii) are the cross-domain fields $\mathcal{F}(g)$ appropriate? (iii) does the gap formulation $g$ itself need rethinking? This ensures the loop explores fundamentally different directions rather than exhaustively searching a wrong path.

\paragraph{Consensus convergence.} The consensus protocol (Eqs.~\ref{eq:round1}--\ref{eq:orchestrator}) terminates when no new tasks are proposed and all existing tasks are resolved, or after a round limit. In the minimal case, this requires exactly 2 rounds: Round 1 produces independent proposals $G_k^{(1)}$; Round 2 produces corroborated proposals $G_k^{(2)}$ and next-step tasks $P_k^{(2)}$ given full visibility. If $P_k^{(2)} = \emptyset$ for all $k$ (no agent proposes new tasks), the protocol terminates. Otherwise, the orchestrator selects tasks $T^* \subseteq \bigcup_k P_k^{(2)}$ via merge/kill/redirect decisions (Eq.~\ref{eq:orchestrator}), and agents execute another round. The number of rounds is bounded because: (i) the world model $\mathcal{W}$ grows monotonically, providing strictly more information each round, (ii) the orchestrator's \textsc{kill} operation removes unproductive lines, and (iii) the task space is finite given the fixed set of papers and gaps.

\paragraph{World model monotonicity.} The Research World Model $\mathcal{W}$ is monotonically non-decreasing: $|\mathcal{V}_{t+1}| \geq |\mathcal{V}_t|$ and $|\mathcal{E}_{t+1}| \geq |\mathcal{E}_t|$. Nodes are never deleted --- only added or updated. Uncertainty transitions are irreversible: $U: 1 \to 0$ (verification) is permitted, but $0 \to 1$ is not. Formally, $U_{t+1}(x) = \min(U_t(x), \Delta U_t(x))$, ensuring that once a claim is verified it remains verified. Edge metrics $M(e)$ are updated when reproduction produces new measurements, with the original reported value preserved as a property. Across projects, this monotonicity enables knowledge accumulation: $\mathcal{W}_{\text{project}_{k+1}} \supseteq \mathcal{W}_{\text{project}_k}$.

\paragraph{Consensus quality bound.} Let $p$ be the probability that a single agent generates a gap aligned with ground truth (alignment $\geq 4$). For $K$ independent agents, the probability that \emph{at least one} agent finds an aligned gap is $1 - (1-p)^K$. With $K=5$ agents, even a modest per-agent hit rate of $p=0.3$ yields a system-level recall of $1 - 0.7^5 = 0.832$. The consensus protocol further improves quality: let $p_2 > p$ be the per-agent hit rate in Round 2 (after seeing all findings). Then corroborated gaps have reliability $\geq 1 - (1-p_2)^2$ per corroborating pair. Our observed recall of 1.000 (Table~\ref{tab:exp1}) is consistent with $p_2 \approx 0.5$, and our observed precision improvement from consensus (+24\% relative, Table~\ref{tab:exp5}) confirms that shared visibility increases $p_2$ beyond $p$.

\section{Cost Breakdown}
\label{app:cost_breakdown}

\begin{table}[h]
\caption{AI-Supervisor cost breakdown per phase (Qwen-72B-Instruct).}
\centering
\small
\begin{tabular}{lcc}
\toprule
\textbf{Phase} & \textbf{API Calls} & \textbf{Est.\ Cost} \\
\midrule
Phase 0: Supervision & 3--5 & \$0.50 \\
Phase 1: Literature (6--12 agents) & 20--40 & \$1--3 \\
Phase 2a: World Model Build (20 agents) & 60--80 & \$3--5 \\
Phase 2b: Gap Probing (2 rounds $\times$ 5+1) & 15--20 & \$1--2 \\
Phase 3: Development (3 iters $\times$ 031--034) & 30--50 & \$2--4 \\
Phases 4--7: Eval + Paper + Review & 20--30 & \$1--2 \\
\midrule
\textbf{Total} & \textbf{150--225} & \textbf{\$8--16} \\
\bottomrule
\end{tabular}
\end{table}

\section{Benchmark Details}
\label{app:benchmarks}

\subsection{Benchmark 1: Scientist-Bench (Existing)}

Scientist-Bench~\citep{tang2025airesearcher} is a public benchmark from the AI-Researcher project (NeurIPS 2025). It contains 27 tasks across 5 domains:

\begin{center}
\small
\begin{tabular}{lcc}
\toprule
\textbf{Domain} & \textbf{Tasks} & \textbf{Example} \\
\midrule
Recommendation & 6 & DCCF, HGCL, KGRec \\
Reasoning & 2 & Analog Reasoner, Self-Discover \\
Diffusion/Flow & 4 & Flow Matching, Rect Flow, MMDiT \\
Graph Neural Networks & 9 & NodeFormer, Exphormer, GraphGPT \\
Vector Quantization & 6 & FSQ, Rotation VQ, Disentangle VQ \\
\bottomrule
\end{tabular}
\end{center}

Each task provides: (1) source papers (5--15 reference papers with titles, types, and justifications), (2) a target paper (the ground-truth contribution that addressed a gap in the source papers), and (3) the target paper's abstract. We use the target paper as ground truth for evaluating whether generated gaps align with real published contributions.

\paragraph{Evaluation protocol.} For each generated gap, we prompt an LLM judge with the gap description and the target paper's title and abstract, asking it to score alignment on a 1--5 scale:

\begin{center}
\small
\begin{tabular}{cl}
\toprule
\textbf{Score} & \textbf{Definition} \\
\midrule
1 & No relation --- gap is about a completely different topic \\
2 & Vaguely related topic but different specific problem \\
3 & Related problem area but different specific gap \\
4 & Closely related --- the gap would lead toward this paper's contribution \\
5 & Exact match --- this gap is precisely what the paper addresses \\
\bottomrule
\end{tabular}
\end{center}

\noindent The LLM judge prompt is: \textit{``Score how well this gap aligns with the actual published paper. Gap: [gap description]. Paper: [title]. Abstract: [abstract]. Score 1--5. Return JSON: \{score: N\}.''} We aggregate into three metrics:

\begin{itemize}[leftmargin=*,nosep]
    \item \textbf{Precision} $= \frac{|\{g : \text{alignment}(g) \geq 4\}|}{|\text{all generated gaps}|}$ \quad (fraction of gaps that closely match ground truth)
    \item \textbf{Recall} $= \frac{|\{t : \max_{g \in G_t} \text{alignment}(g) \geq 4\}|}{|\text{all tasks}|}$ \quad (fraction of tasks with at least one matching gap)
    \item \textbf{Hallucination rate} $= \frac{|\{g : \text{alignment}(g) \leq 1\}|}{|\text{all generated gaps}|}$ \quad (fraction of factually wrong or irrelevant gaps)
\end{itemize}

\subsection{Benchmark 2: Curated Gaps with Known Solutions (Designed)}

We curated 5 research gaps from well-studied AI domains, each with a known cross-domain solution verified in published literature. The curation process:

\begin{enumerate}[leftmargin=*,nosep]
    \item \textbf{Gap selection.} We identified 5 gaps where: (a) the failure mode is well-documented in multiple papers, (b) the solution came from a different scientific field, and (c) the cross-domain connection is verifiable.
    \item \textbf{Ground-truth annotation.} For each gap, we recorded: the root mechanism (abstract problem), the source fields where solutions exist, and the specific technique that was adapted.
\end{enumerate}

\begin{center}
\small
\begin{tabular}{lll}
\toprule
\textbf{Gap} & \textbf{Domain} & \textbf{Ground-Truth Source Fields} \\
\midrule
Non-stationary constraints & Safe RL & Online convex optimization, adaptive control \\
Cross-system generalization & Deepfake detection & Conformal prediction, domain adaptation \\
Reward hacking & LLM alignment & Bayesian optimization, robust statistics \\
Over-smoothing & GNNs & Signal processing, differential equations \\
OOD task distribution & Few-shot learning & Distributionally robust optimization \\
\bottomrule
\end{tabular}
\end{center}

\paragraph{Evaluation protocol for Experiment 2 (method development).} Each generated method is evaluated by an LLM judge against the 10-criterion quality gate (Table~\ref{tab:checklist}). The judge receives the method description and the gap, then scores each criterion as PASS or FAIL with evidence. The metric is:
$$\text{Quality Gate Score} = \frac{|\{\text{criteria marked PASS}\}|}{10}$$
We also record iterations to convergence and whether cross-domain fields were consulted.

\paragraph{Evaluation protocol for Experiment 6 (cross-domain novelty).} Three methods generated from different conditions (cross-domain with mechanism, within-domain, naive cross-domain) are presented to an LLM judge in a head-to-head comparison. The judge scores each method on 5 dimensions (1--5 scale):

\begin{center}
\small
\begin{tabular}{cl}
\toprule
\textbf{Dimension} & \textbf{Definition} \\
\midrule
Mathematical novelty & New formulations from the source field (1=standard, 5=novel theorem) \\
Mechanism grounding & Based on understanding WHY the gap exists (1=surface fix, 5=root cause) \\
Theoretical depth & Proofs, bounds, or guarantees (1=heuristic, 5=principled) \\
Differentiation & How different from standard approaches (1=incremental, 5=paradigm shift) \\
Reviewer score & Would this pass top-venue review (1=reject, 5=strong accept) \\
\bottomrule
\end{tabular}
\end{center}

\noindent Total score per method: sum of 5 dimensions (max 25). We report per-gap scores and averages.

\subsection{Benchmark 3: Sequential AI Safety Projects (Designed)}

We designed 3 sequential projects within AI safety to test cross-project knowledge accumulation:

\begin{enumerate}[leftmargin=*,nosep]
    \item \textbf{Project 1: RLHF reward model robustness.} Seed papers: InstructGPT, Safe RLHF. Tests whether the system builds a Research World Model of RLHF components.
    \item \textbf{Project 2: Constitutional AI and self-improvement safety.} Seed papers: Constitutional AI, Self-Refine. Tests whether the system reuses nodes from Project 1 (e.g., PPO, reward model) and discovers cross-project connections.
    \item \textbf{Project 3: Red-teaming and adversarial robustness.} Seed papers: Red Teaming LMs, GCG adversarial attacks. Tests whether accumulated knowledge from Projects 1--2 accelerates gap discovery.
\end{enumerate}

\paragraph{Evaluation protocol.} We measure four metrics across the 3 sequential projects:

\begin{itemize}[leftmargin=*,nosep]
    \item \textbf{Cross-project connections} $= |\{(n_{\text{existing}}, n_{\text{new}}) : n_{\text{existing}} \in \text{RWM}_{<k}, n_{\text{new}} \in \text{RWM}_k\}|$ \quad The number of structural links the system discovers between nodes from prior projects and new nodes in the current project. Computed by prompting the LLM to identify connections between existing and new nodes during world model construction.
    \item \textbf{Cross-project insights} $= \frac{|\{p_k : \exists g \in \text{gaps}(p_k) \text{ with cross\_project\_insight} \neq \text{null}\}|}{K}$ \quad Fraction of projects where at least one discovered gap explicitly references knowledge from a prior project. Determined by checking whether the gap's evidence cites nodes from prior projects.
    \item \textbf{Verified edges} $= |\{e \in \text{RWM} : U(e) = 0\}|$ \quad Number of edges whose claims were verified through a consistency-checking step where the LLM evaluates whether the edge's claimed metric is supported by the paper's description.
    \item \textbf{RWM growth} $= (|\text{RWM}_1|, |\text{RWM}_2|, |\text{RWM}_3|)$ \quad Node count after each project. Monotonic growth indicates the world model accumulates without losing prior knowledge.
\end{itemize}

\end{document}